\documentclass[letterpaper, 10 pt, conference]{ieeeconf}  % Comment this line out if you need a4paper

\IEEEoverridecommandlockouts                              % This command is only needed if 
\usepackage{balance} % for balancing columns on the final page
\usepackage[english]{babel}
\usepackage{stfloats}
\usepackage{graphicx}
\usepackage{subfigure}

\graphicspath{ {figures/} }
\usepackage{makecell}
\usepackage{sidecap}
\usepackage{amsmath}
\usepackage{multirow}
\usepackage{amssymb}
\usepackage{hyperref}

\usepackage{algorithm}
\usepackage{algorithmicx}
\usepackage{algpseudocode}
\usepackage{epstopdf}
\usepackage{booktabs}
\usepackage{wrapfig}
\usepackage{marvosym}

\title{\LARGE \bf
Intent-based Deep Reinforcement Learning for Multi-agent Informative Path Planning
}

\author{Tianze Yang$^{1}$, Yuhong Cao$^{1}$, Guillaume Sartoretti$^{1}$$^{\dagger}$% <-this % stops a space
\thanks{$^{\dagger}$ Corresponding author, to whom correspondence should be addressed}% <-this % stops a space
\thanks{$^{1}$Authors are with the Department of Mechanical Engineering, College of Design and Engineering, National University of Singapore.
        {\tt\small yangtianze@u.nus.edu, caoyuhong@u.nus.edu, mpegas@nus.edu.sg}}%
}

\begin{document}

\maketitle
\thispagestyle{empty}
\pagestyle{empty}

%%%%%%%%%%%%%%%%%%%%%%%%%%%%%%%%%%%%%%%%%%%%%%%%%%%%%%%%%%%%%%%%%%%%%%%%%%%%%%%%
%%%%%%%%%%%%%%%%%%%%%%%%%%%%%%%%%%%%%%%%%%%%%%%%%%%%%%%%%%%%%%%%%%%%%%%%%%%%%%%%

\begin{abstract}

In multi-agent informative path planning (MAIPP), agents must collectively construct a global \textit{belief} map of an underlying distribution of interest (e.g., gas concentration, light intensity, or pollution levels) over a given domain, based on measurements taken along their trajectory. They must frequently replan their path to balance the exploration of new areas with the exploitation of known high-interest areas, to maximize information gain within a predefined budget. Traditional approaches rely on reactive path planning conditioned on other agents' predicted future actions. However, as the belief is continuously updated, the predicted actions may not match the executed actions, introducing noise and reducing performance. We propose a decentralized, deep reinforcement learning (DRL) approach using an attention-based neural network, where agents optimize long-term individual and cooperative objectives by sharing their \textit{intent}, represented as a distribution of medium-/long-term future positions obtained from their own policy. Intent sharing enables agents to learn to claim or avoid broader areas, while the use of attention mechanisms allows them to identify useful portions of imperfect predictions, maximizing cooperation even based on imperfect information. Our experiments compare the performance of our approach, its variants, and high-quality baselines across various MAIPP scenarios. We finally demonstrate the effectiveness of our approach under limited communication ranges, towards deployments under realistic communication constraints.

\end{abstract}

%%%%%%%%%%%%%%%%%%%%%%%%%%%%%%%%%%%%%%%%%%%%%%%%%%%%%%%%%%%%%%%%%%%%%%%%%%%%%%%%
%%%%%%%%%%%%%%%%%%%%%%%%%%%%%%%%%%%%%%%%%%%%%%%%%%%%%%%%%%%%%%%%%%%%%%%%%%%%%%%%

\section{INTRODUCTION}
Many robotic deployments rely on information gathering, such as autonomous exploration~\cite{mascarich2018multi}, environment monitoring~\cite{dunbabin2012robots}, search-and-rescue~\cite{baxter2007multi}, and surface inspection~\cite{wu2015path}.
Starting from a (usually empty) initial belief of the environment, the agent needs to reason about known, informative areas as well as unknown ones to plan an informative path and collect information to update its belief under a given budget constraint (e.g., maximum path length or mission duration).
Such deployments are known as adaptive \textit{informative path planning} (IPP)~\cite{lim2016adaptive,schmid2020efficient}.
 
\begin{figure}[tb]
\centering
\subfigure[Ground truth interest map]{
\begin{minipage}[t]{0.5\linewidth}
\centering
\includegraphics[width=3.7cm, height=3.4cm]{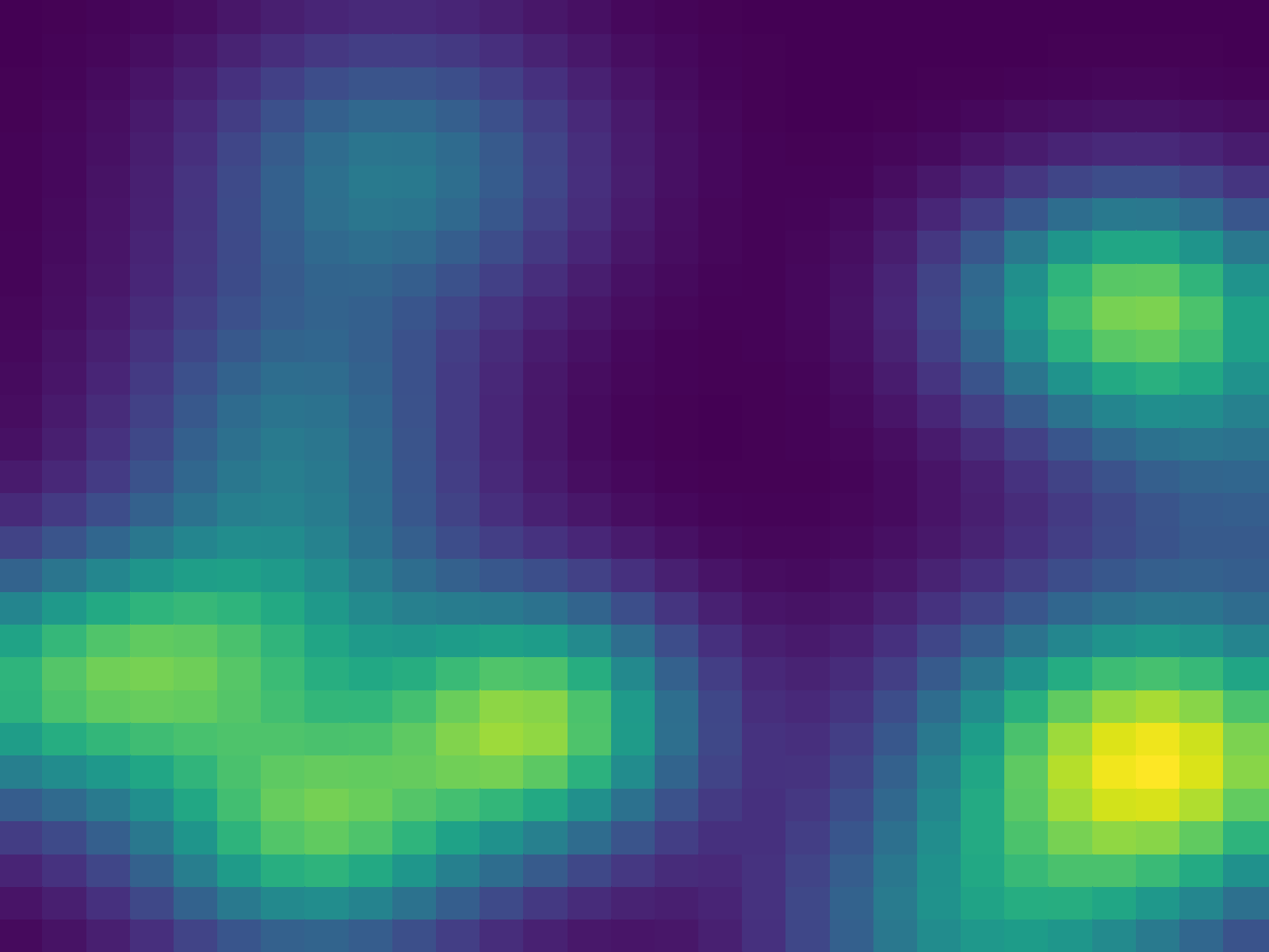}
\end{minipage}%
}%
\subfigure[Predicted interest map]{
\begin{minipage}[t]{0.5\linewidth}
\centering
\includegraphics[width=3.7cm, height=3.4cm]{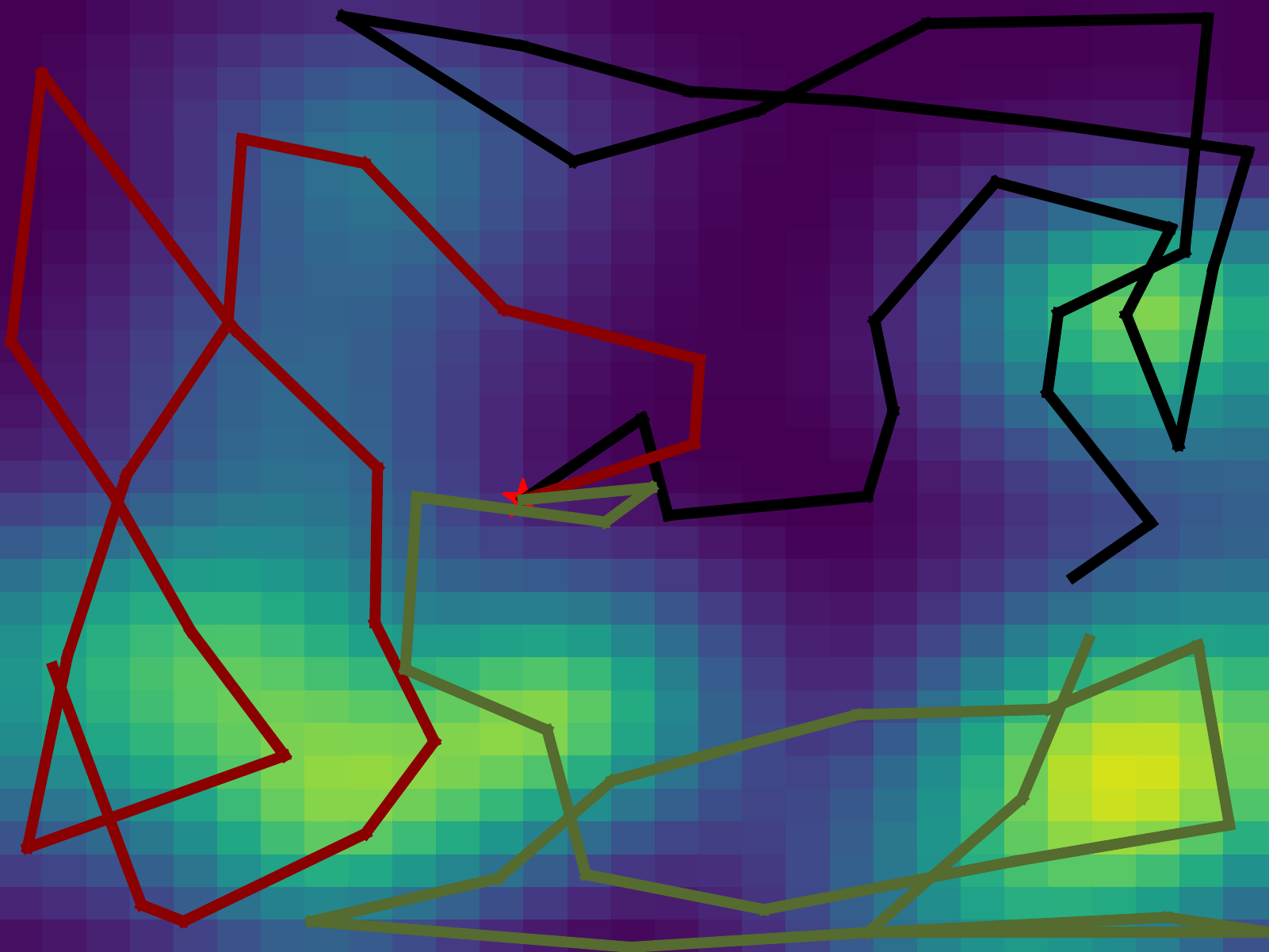}
\end{minipage}
}
\quad

\subfigure[Predicted standard deviation]{
\begin{minipage}[t]{0.5\linewidth}
\centering
\includegraphics[width=3.7cm, height=3.4cm]{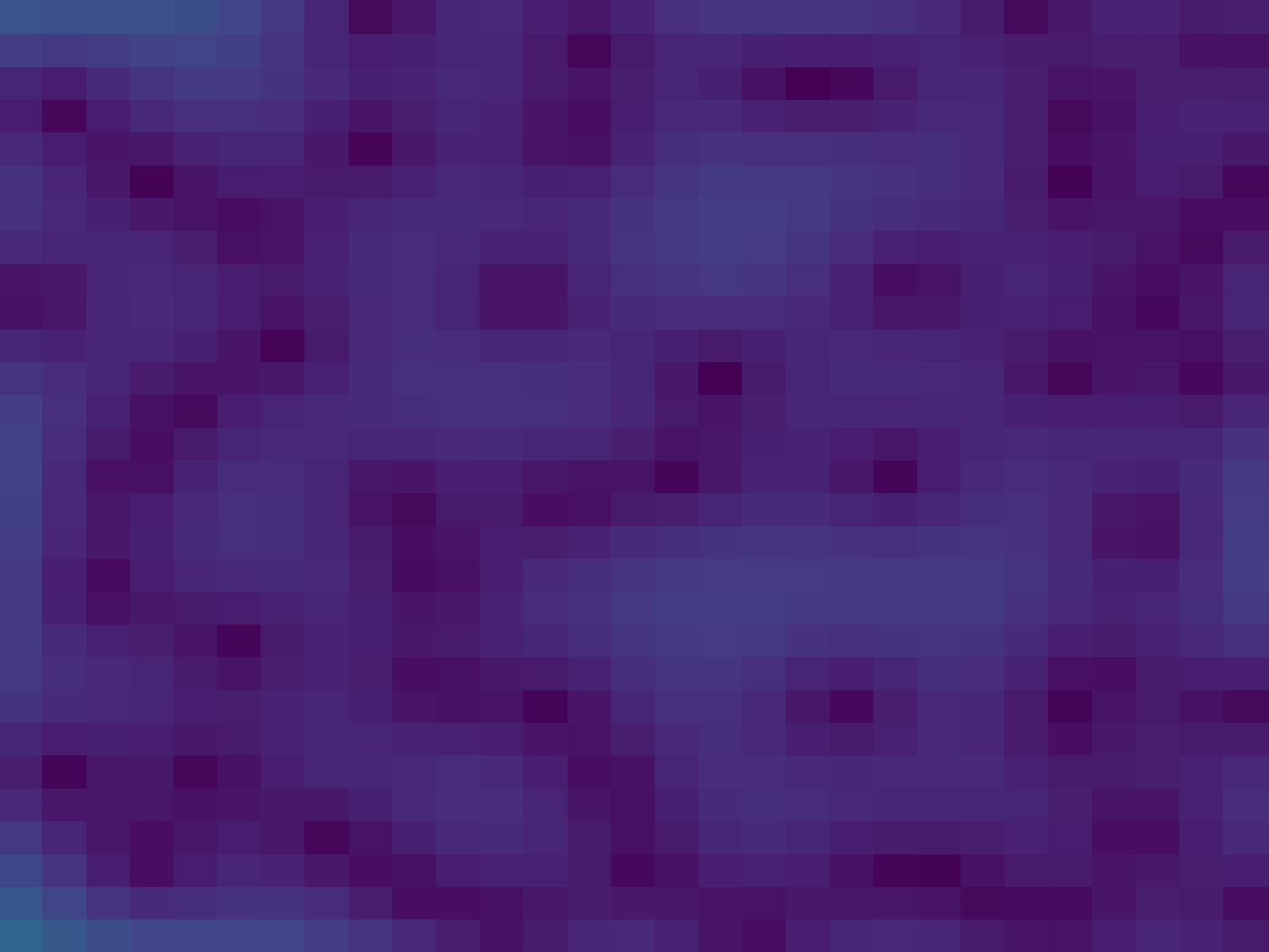}
\end{minipage}%
}%
\subfigure[High interest area]{
\begin{minipage}[t]{0.5\linewidth}
\centering
\includegraphics[width=3.7cm, height=3.4cm]{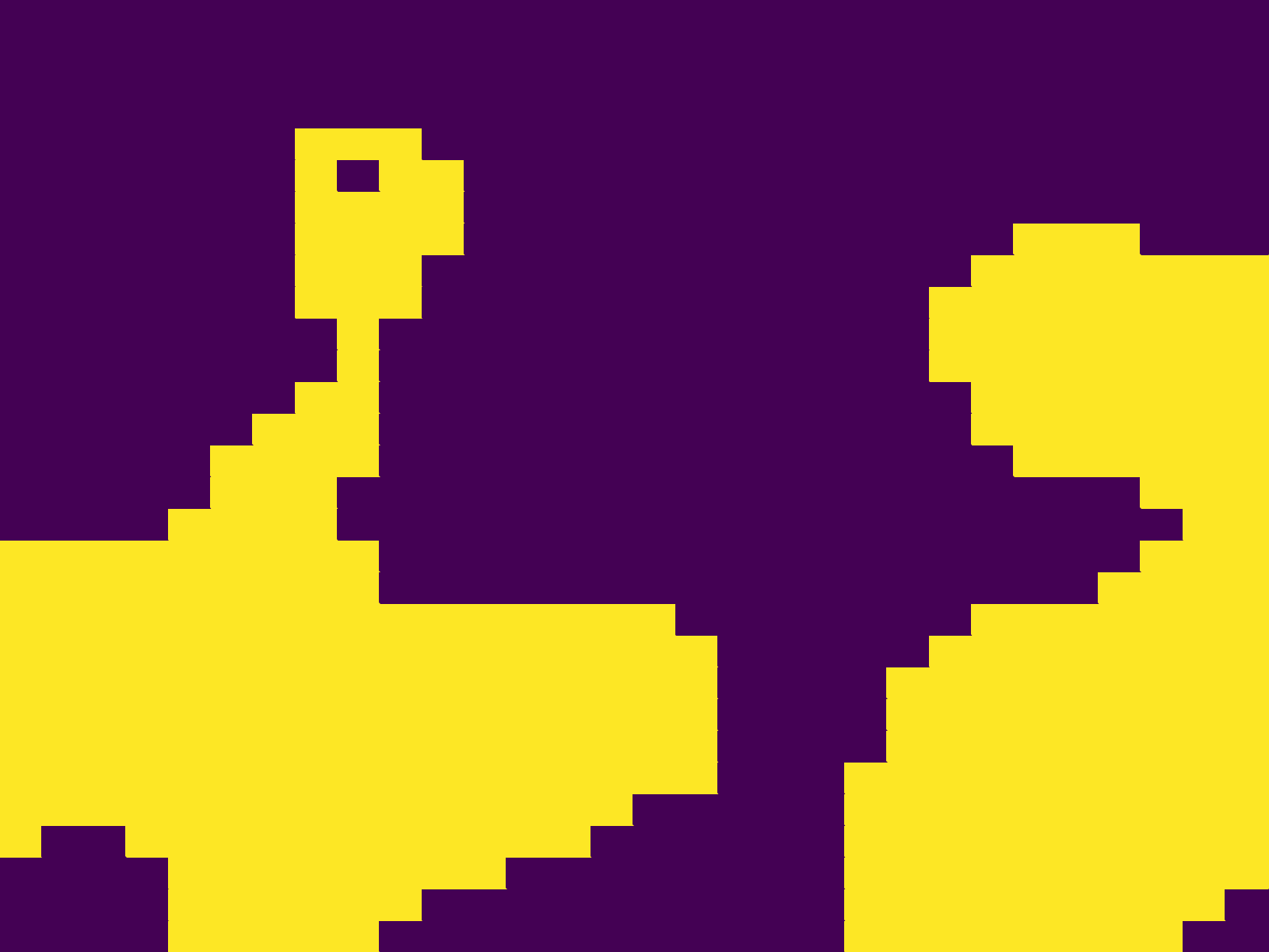}
\end{minipage}
}
% \quad
\vspace{-0.5cm}
\caption{\textbf{Example MAIPP instance with 3 agents}. The black, green, and red lines represent the trajectories of these $3$ agents, seen spreading out initially to distributedly cover the domain, and then regrouping (in particular, the black and green agents) to cover the lower right high-interest area into more details. Note that agents start without any prior knowledge over the ground truth.}
\label{example of MAIPP problem}
\vspace{-0.5cm}
\end{figure}

As a natural extension of such single-agent IPP, the \textit{Multi-Agent Informative Path Planning} (MAIPP) aims to further improve efficiency through distributed cooperation.
There, multiple agents collect information individually but share their measurements in real-time to construct a \textit{global} agents' belief, which estimates the true, underlying interest map from a finite set of discrete measurements.
To achieve good cooperation in MAIPP, agents are expected to make \textit{non-myopic} decisions that balance the trade-off between distributing to explore the environment and regrouping to exploit a known, high-interest area meticulously.
For example, Fig.~\ref{example of MAIPP problem} illustrates an episode of an MAIPP instance, to visualize this trade-off.
Starting from the same position, the three agents start by distributing to cover the environment; later, the black and green agents regroup to collectively cover the lower-right high-interest area in detail.

Cooperation in MAIPP can be achieved through sequential greedy assignment (SGA)~\cite{corah2017efficient}. In SGA, agents plan their paths sequentially, considering the paths planned by all other agents before them. This allows agents to be aware of others' future actions. As the interest map is updated online, agents often plan paths in a receding-horizon manner to remain responsive to map updates. In scenarios where the interest map remains stable, agents can replan long-horizon paths based on existing plans from other agents, leading to efficient long-term cooperation. However, when the interest map changes frequently and drastically, shared paths can become volatile and confuse agents.

To address this issue, we propose a framework that utilizes an attention-based model trained with distributed deep reinforcement learning (DRL). Each agent plans its path based on the global belief and the \textit{intent} of other agents, represented as a spatial distribution of their predicted medium-/long-term positions. This framework facilitates efficient cooperation in MAIPP: at each decision step, agents use their learned policy and knowledge of other agents' intents to plan multiple fixed-horizon trajectories and construct a spatial \textit{intent distribution}. This distribution is then shared with all agents at low communication cost. Each agent combines the intent distributions of other agents into a single map, which serves as an additional input to our model, along with the agent's individual belief. The intent distribution allows agents to learn to claim/avoid areas based on other agents' ongoing strategies. Furthermore, our attention-based network enables agents to recognize the useful portions of imperfect intent, unlike conventional methods that assume perfect intent and struggle with frequent plan changes.

Through numerical experiments, we demonstrate the effectiveness of our approach compared to two SGA-based baselines. We highlight the importance and performance benefits of each technique in our framework. Additionally, we evaluate our models (trained under global communications) under limited communication ranges, simulating practical deployments (where global communication may not be a practical/feasible assumption). Remarkably, our learning-based approach maintains performance without requiring retraining or fine-tuning, showcasing its potential for deployments under realistic communication constraints.

%%%%%%%%%%%%%%%%%%%%%%%%%%%%%%%%%%%%%%%%%%%%%%%%%%%%%%%%%%%%%%%%%%%%%%%%%%%%%%%%
%%%%%%%%%%%%%%%%%%%%%%%%%%%%%%%%%%%%%%%%%%%%%%%%%%%%%%%%%%%%%%%%%%%%%%%%%%%%%%%%

\section{RELATED WORKS}
\label{Related works}

Single-agent IPP has drawn the community's attention for more than ten years~\cite{marchant2014bayesian, binney2012branch}.
Recent works have mainly focused on adaptive IPP, where the agent starts with an empty initial belief and where frequent replanning is necessary as this belief is updated online with new measurements.
There, the real-time requirements drastically lead state-of-the-art IPP planners to focus on cutting down planning times through meta-heuristic and deep reinforcement learning~\cite{hitz2017adaptive,popovic2020informative, cao2022catnipp, ruckin2022adaptive}.
Notably, our recent work~\cite{cao2022catnipp} recently proposed a DRL-based single-agent IPP approach that learns to utilize the full agent's belief to make reactive, yet non-myopic decisions.
Multi-agent IPP has been relatively less studied so far.
Most relevant, Viseras et al.~\cite{viseras2016decentralized} proposed to approach multi-agent IPP by combining a greedy strategy and collision avoidance.
However, lessons from single-agent IPP planners have shown that greedy strategies will often lead to myopic decisions and thus poorer long-term efficiency~\cite{hollinger2014sampling,hitz2017adaptive,arora2017randomized,popovic2020informative}.
This issue is further aggravated in a cooperative multi-agent system.
A promising way to extend the recent successes in single-agent IPP to multi-agent IPP (in particular, in a distributed manner) is through sequential greedy assignment (SGA)~\cite{corah2017efficient}, which has been designed and validated in other information-gathering scenarios.
There, each agent sequentially plans its own individual path, conditioned on all paths planned by other agents before itself, thus leading to scalable, priority-based cooperation.
SGA can essentially be implemented on top of any conventional single-agent IPP planner,
and the final performance will ultimately depend on the choice of this underlying planner~\cite{corah2017efficient}.

%%%%%%%%%%%%%%%%%%%%%%%%%%%%%%%%%%%%%%%%%%%%%%%%%%%%%%%%%%%%%%%%%%%%%%%%%%%%%%%%
%%%%%%%%%%%%%%%%%%%%%%%%%%%%%%%%%%%%%%%%%%%%%%%%%%%%%%%%%%%%%%%%%%%%%%%%%%%%%%%%

\section{BACKGROUND}
\label{Preliminaries}

Following~\cite{marchant2014bayesian, viseras2016decentralized, mishra2018online, hitz2017adaptive, cao2022catnipp}, in this work, we rely on Gaussian Processes (GPs) to construct and update the agents' beliefs and formulate adaptive MAIPP based on GPs.

%%%%%%%%%%%%%%%%%%%%%%%%%%%%%%%%%%%%%%%%%%%%%%%%%%%%%%%%%%%%%%%%%%%%%%%%%%%%%%%%

\subsection{Gaussian Process} 

In IPP, the underlying distribution of \emph{interest} over the 2D environment $\mathcal{E} \subset \mathbb{R}^2$ is typically modeled as a continuous function $\zeta: \mathcal{E}  \xrightarrow{} \mathbb{R}$ (and is not available to the agent(s)).
Gaussian Processes offer us a natural means to interpolate between discrete measurements, in order to build an approximation $\mathcal{GP}(\mu, P) \approx \zeta$ of this ground truth.
Given a set of $n'$ locations $\mathcal{X}^* \subset \mathcal{E}$ where the interest is to be inferred, a set of $n$ measured locations $\mathcal{X} \subset \mathcal{E}$ and the sampled measurement set $\mathcal{Y}$, the mean and covariance of the GP are obtained as:
\begin{equation}
\mu = \mu(\mathcal{X}^*)+K(\mathcal{X}^*, \mathcal{X})[K(\mathcal{X},\mathcal{X})+\sigma_n^2I]^{-1}(\mathcal{Y}-\mu(\mathcal{X})), \nonumber    
\end{equation}
\begin{equation}
\resizebox{0.48\textwidth}{!}{
$P = K(\mathcal{X}^*, \mathcal{X}^*)-K(\mathcal{X}^*, \mathcal{X})[K(\mathcal{X},\mathcal{X})+\sigma_n^2I]^{-1}\times K(\mathcal{X}^*,\mathcal{X})^T \nonumber$}
\end{equation}
where $K(\cdot)$ represents a pre-defined kernel function, $\sigma^2_n$ is a parameter representing the measurement noise, and $I$ is a $n\times n$ identity matrix.
Specifically, our work uses the Matérn $3/2$ kernel following ~\cite{popovic2020informative,ghaffari2019sampling}.

%%%%%%%%%%%%%%%%%%%%%%%%%%%%%%%%%%%%%%%%%%%%%%%%%%%%%%%%%%%%%%%%%%%%%%%%%%%%%%%%

\subsection{Multi-Agent Adaptive Informative Path Planning}

A solution to an MAIPP instance is a set of agent trajectories $\psi = \left\{\psi^1,...,\psi^m \right\}$.
MAIPP seeks an optimal trajectory set $\psi^*$ from the set of all available trajectories $\Psi$, which maximizes the information gain from all measurements obtained from all agent trajectories while satisfying some budget constraint for each agent:
\vspace{-0.2cm}
\begin{equation}
    \psi^\ast = \mathop{{\rm {argmax}}} \limits_{\psi \in \Psi}\sum_{i=1}^{m} {\rm I}(\psi^i),\ {\rm {s.t.\ C}} (\psi^i) \leq B, 1\leq i\leq m,
    \label{eq:obj}
\end{equation}
where ${\rm I}$ denotes the information gained by agent $i$, $\rm{C}(\psi^i)$ the trajectory length of agent $i$, $B$ the individual budget (i.e., max trajectory length) for each agent (here, assumed identical for all agents, but this assumption can be lifted).

In line with recent works~\cite{popovic2020informative,cao2022catnipp}, in this work, we define the information gain as the uncertainty reduction over the high-interest areas: ${\rm I}(\psi^i) = {\rm Tr}(P_I^-)-Tr(P_I^+)$, where $\rm Tr(\cdot)$ denotes the trace of a matrix, and $P_I^-$ and $P_I^+$ are the prior and posterior covariance of the high-interest areas.
These high-interest areas are defined by an \textit{upper confidence bound}: $\mathcal{X}_I=\left\{x_i \in \mathcal{X}^* \mid \mu_i^{-}+\beta P_{i, i}^{-} \geq \mu_{t h}\right\}$, where $\mu_i^{-}$ and $P_{i, i}^{-}$ are the mean and variance of the GP at the location $x_{i}$, and $\mu_{t h}, \beta \in \mathbb{R}^{+}$ control the threshold and confidence interval, respectively.
That is, by replacing $\mathcal{X}^*$ with $\mathcal{X}_I$ when calculating the covariance, the information gain in the objective function Eq.\eqref{eq:obj} is restricted to the high-interest areas predicted by the GP.
Consequently, the planner needs to be reactive and adaptive to the change of predicted high-interest areas, which are updated along with new measurements.

%%%%%%%%%%%%%%%%%%%%%%%%%%%%%%%%%%%%%%%%%%%%%%%%%%%%%%%%%%%%%%%%%%%%%%%%%%%%%%%%%%%%%%%%%%%%%%%%%%%%%%%%%%%%%%%%%%%%%%%%%%%%%%%%%%%%%%%%%%%%%%%%%%%%%%%%%%%%%%%%

\section{METHOD}

%%%%%%%%%%%%%%%%%%%%%%%%%%%%%%%%%%%%%%%%%%%%%%%%%%%%%%%%%%%%%%%%%%%%%%%%%%%%%%%%

\subsection{MAIPP as a Sequential Decision-making problem}

We formulate MAIPP as a sequential decision-making problem on graphs to decrease the complexity of a continuous environment.
Each agent $i$ maintains its individual \textit{waypoint graph} $G^i$, which is constructed by a probabilistic roadmap (PRM)~\cite{latombe1998probabilistic} at the beginning of the mission and covers the whole search domain.
Specifically, $n$ nodes are randomly sampled and connected to their $k$ nearest neighbors, such that $G^i=(V^i,E^i), i\in \left\{1,...,m\right\}$, where $V^i=\left\{v^i_1,...,v^1_n\right\}$ is a set of nodes and $E^i$ a set of edges.
We do not assume the presence of any obstacles over the domain, but note that adding obstacles would be easy, as it would only require pruning any edge that collides with an obstacle in each agent's graph (similar to~\cite{cao2023ariadne}).

All agents start from the same node at the beginning of the mission.
At each of its decision step (i.e., upon reaching its next waypoint), an agent selects one of its neighboring nodes as its next waypoint according to its individual policy $p$ and immediately sets course for it along a straight line and at a constant speed.
Note that this parallel decision-making is done asynchronously, since agents typically reach their next waypoints at non-synchronous times during the mission.
Agents continue constructing their path iteratively until their budget is exhausted, constructing the final trajectory $\psi^i=\{\psi^i_1,...,\psi^i_c\}$, $\forall \psi^i_n\in V^i, (\psi^i_j, \psi^i_{j+1})\in E^i$, with trajectory length $C(\psi^i_c)=\sum \limits_{j=1}^{n_i-1}L_2(\psi^i_j,\psi^i_{j+1})$, where $L_2(\cdot,\cdot)$ denotes the Euclidean distance between two nodes.  
As a result, in our work, the MAIPP problem can be seen as a set of cooperative, sequential decision-making problems on a set of parallel graphs $G=\left\{G^1, G^2,..., G^m \right\}$.

%%%%%%%%%%%%%%%%%%%%%%%%%%%%%%%%%%%%%%%%%%%%%%%%%%%%%%%%%%%%%%%%%%%%%%%%%%%%%%%%

\subsection{Agent's Intent}

We term the probabilistic distribution of a given agent's future positions as the \textit{intent} of that agent.
Each agent will often update and share its most recent intent and receive that of all other agents, which it will fuse into an accumulated intent map to use in its own decision-making.
In our approach, each agent generates its own intent by sampling trajectories from its individual policy, which is conditioned on other agents' most recent intent. At each decision step, agent $i$ plans multiple trajectories using its policy $p$. During this predictive sampling process, agent $i$ virtually updates its position and the covariance matrix output by the Gaussian process (GP), assuming other agents are stationary. From these sampled predicted trajectories, agent $i$ fits a Gaussian distribution $GD(\mu^i(t), \Sigma^i(t))$ over some (or all, see Sect.~\ref{Experiments} for more details) of the nodes from these sampled predicted trajectories, with mean and covariance $\mu^i(t)$ and $\Sigma^i(t)$. We also set a minimum bound for the resulting covariance matrix to avoid singularities.

Agent $i$ updates its own interest map after reaching each node along its path. It receives updated intent maps from the other $m-1$ agents and fuses them into an accumulated intent distribution by summing the corresponding $m-1$ Gaussian distributions and normalizing the result. Agent $i$ uses its own belief and accumulated intent distribution to sample future paths, update its intent distribution, and select its next action. It then broadcasts its new $\mu^i(t)$ and $\Sigma^i(t)$ (describing its updated intent distribution) to all other agents. It is important to note that agent $i$ only needs to broadcast a vector $\left [\mu^i(t), \Sigma^i(t) \right ]$ to other agents, and the Gaussian distribution $GD(\mu^i(t), \Sigma^i(t))$ is constructed by other agents when they need to make a decision. This low communication cost allows for scalability and real-life implementation.

%%%%%%%%%%%%%%%%%%%%%%%%%%%%%%%%%%%%%%%%%%%%%%%%%%%%%%%%%%%%%%%%%%%%%%%%%%%%%%%%

\subsection{RL Cast}

\begin{figure*}[t]
    \vspace{0.2cm}
    \centering
    \includegraphics[width=0.94\linewidth]{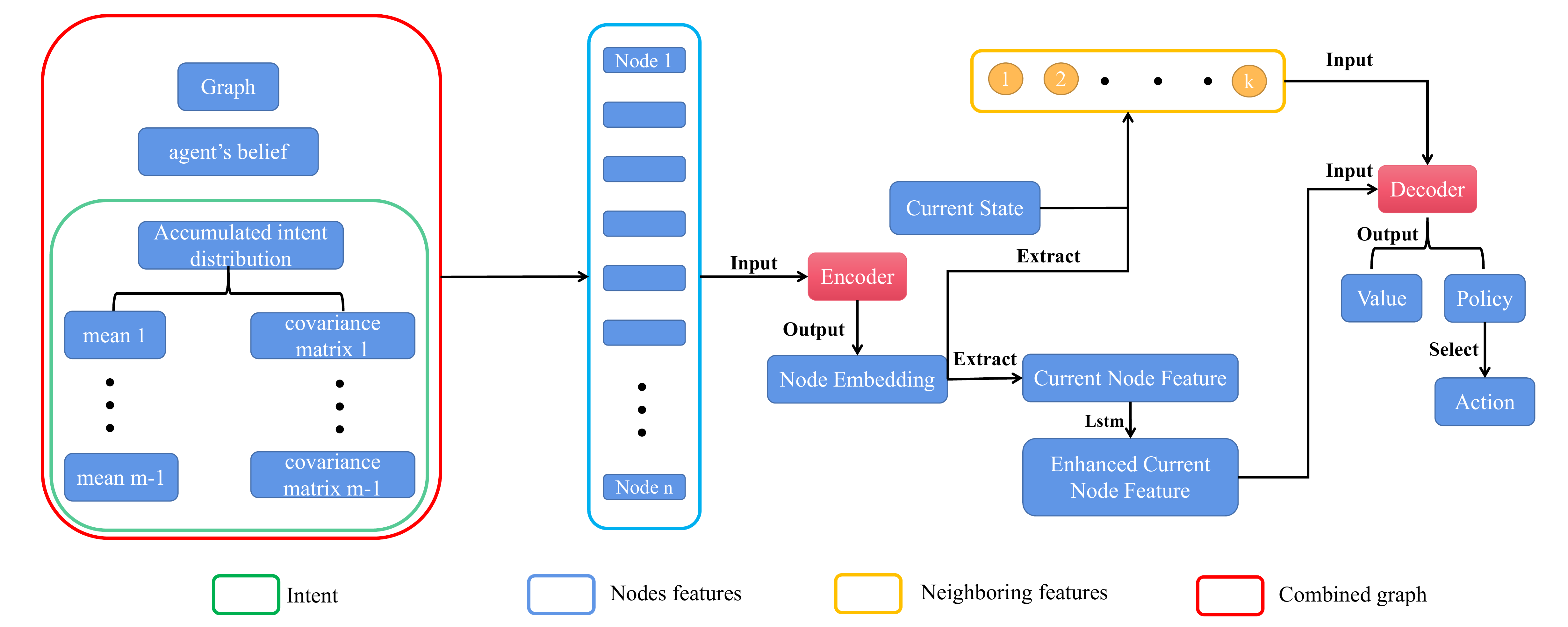}
    \vspace{-0.45cm}
    \caption{\textbf{Attention-based neural network structure}. The (augmented) nodes of the combined graph (circled in red) are the first input to the network, and contain information about: 1) their location, 2) the agents' belief (predicted by the GP) at their location, and 3) the intent level, obtained from the accumulated intent distribution (received from the other $m-1$ agents) at their location. Relying on the self-attention mechanism in the encoder, each node reasons about the dependencies between all other nodes and itself (i.e., learning the relationship between each node position, the global belief, and the intent distribution). After that, the decoder takes the output of this encoder (the \textit{nodes embedding}), and combines it with the current state (agent's position, remaining budget, and trajectory executed so far), to finally select one of the neighboring nodes as the next position to visit.}
    \label{figure1}
    \vspace{-0.3cm}
\end{figure*}

\textbf{Observation}:
The observation for agent $i$ at decision step $t$ is $s^i_t=\{G^{'i},v^i_c,B^i_c,\psi^i\}$, which consists of the combined graph and the current state.
In MAIPP, the combined graph $G^{'i}=(V^{'i}, E^i)$ integrates information from the agent's own waypoint graph, the current agent's belief predicted by the GP, and the intent of other $m-1$ agents at each discrete node location: specifically, each node $v^{'i}_j=(x^i_j, y^i_j, \mu(v^i_j), P(v^i_j),f(v^i_j)) \in V^{'i}$.
The current state of the agent is $\{v^i_c,B^i_c,\psi^i\}$, where $v^i_c=(x^i_c,y^i_c) \in V^i$ is the current position of the agent $i$, $\psi^i=(\psi^i_1,...,\psi^i_c)$ is the agent's executed trajectory, and $B^i_c=B-{\rm C}(\psi^i)$ its remaining budget.

\textbf{Action}: When agent $i$ arrives at its most recently selected node, its belief is updated according to all measurements taken by all $m$ agents since agent $i$ made its last decision (i.e., at its previous node).
The intent is also updated based on all other agents' future predictions made since then.
Then, based on this updated information, the agent's attention-based neural network outputs a policy over its neighboring nodes.
Finally, agent $i$ selects an action, i.e., the next node to visit from its policy, and starts moving in a straight line to that next waypoint.

\textbf{Reward}: At decision step $t$, agent $i$ is given a positive reward according to its most recent action, proportional to the amount of uncertainty reduction: $r_t=({\rm Tr}(P^{t-1})-{\rm Tr}(P^{t}))/{{\rm Tr}(P^{t-1})}$, ${\rm Tr}(P^{t-1})$ is used to scale the reward.
Note that other agents may have also made decisions at step $t-1$ that can have influenced this uncertainty reduction.
Second, a final negative reward $r_f=-\beta\cdot{\rm Tr}(P^{f})$ is assigned to agent $i$'s when its budget is exhausted, where $P^f$ is the remaining covariance in high-interest areas, and $\beta$ is a scaling factor ($0.02$ in practice).

%%%%%%%%%%%%%%%%%%%%%%%%%%%%%%%%%%%%%%%%%%%%%%%%%%%%%%%%%%%%%%%%%%%%%%%%%%%%%%%%

\subsection{Neural Network}

Inspired by the success of~\cite{cao2022catnipp} in single agent IPP, we rely on attention layers~\cite{Vaswaniattention} as the fundamental modules of our neural network for their powerful ability to capture dependencies between high-interest areas at multiple scales.
Our network consists of two parts: an encoder that learns to reason about the relationships over the agent's belief and others' intents between nodes in the waypoint graph, and a decoder that outputs a policy over the next node to visit based on the encoder's output, the agent position, high-interest threshold, remaining budget, and the executed trajectory so~far.

\textbf{Attention Layer}:
The input of an attention layer consists of a query source $h^q$ and a key-and-value source $h^{k,v}$. The main principle for this general (cross-)attention mechanism is:
$
q_{i}=W^Qh^{q}_{i}, \ k_{i}=W^Kh^{k,v}_{i}, \ v_{i}=W^Vh^{k,v}_{i},
    u_{ij}=\frac{q_{i}^T\cdot k_{j}}{\sqrt{d}}, \  a_{ij}=\frac{e^{u_{ij}}}{\sum_{j=1}^{n}e^{u_{ij}}}, \   h'_{i}=\sum_{j=1}^{n}a_{ij}v_{j},
$
Where all weight matrices are learnable.

\textbf{Encoder}
In the encoder, we first project each node in $V^{'i}$ into a higher-dimensional (128 in practice) feature vector, and add a graph positional embedding to each node feature as in~\cite{dwivedi2020generalization} to represent edge connectivity.
We then input these high-dimensional node features as both the query source and key-and-value source for the following attention layer, which is trained to let each node adaptively merge other nodes' features into its own feature vector.
We term the final encoder's output the \textit{Node Embeddings}.

\textbf{Decoder}
In the decoder, we first extract the node embeddings of the robot's current node (termed \textit{current features}) as well as its neighboring node embeddings (termed \textit{neighboring features}).
We then enhance the current features by adding information about the high-interest threshold, remaining budget, and executed trajectory so far.
Specifically, we concatenate the current feature with the high-interest threshold and remaining budget, then project it back to its previous dimension (128).
This updated current feature is then passed through an LSTM~\cite{article} layer to yield the \textit{enhanced current node features}, where the hidden and cell states are input from the previous enhanced current feature along the executed trajectory.
Together with the neighboring features as the Key, we input the enhanced current node features as the Query to a pointer layer~\cite{vinyals2015pointer}, a special attention layer that directly uses normalized attention weights as output.
We regard these attention weights as our final policy $\pi$, a probability distribution over neighboring nodes whose dimensions automatically depend on the number of neighboring nodes.

%%%%%%%%%%%%%%%%%%%%%%%%%%%%%%%%%%%%%%%%%%%%%%%%%%%%%%%%%%%%%%%%%%%%%%%%%%%%%%%%

\subsection{Training}

We train our model using PPO~\cite{DBLP:journals/corr/SchulmanWDRK17}.
At the beginning of each training episode, we randomly generate $8$ to $12$ 2D Gaussian distributions in the $[0,1]^2$ space, to construct the ground truth that agents will explore.
The resolution of the ground truth and the agent's belief is $30 \times 30$, and thus the covariance matrix trace (i.e., the uncertainty over the agent's belief, initially empty), starts at $900$.
The start position of all agents is randomly drawn within $[0,1]^2$.
A PRM randomly samples $200$ nodes to build a graph for each agent, in which we connect each node to its nearest $k=20$ neighbors.
We train with $m=3$ agents under a budget of $B=3$ for each agent.
An agent takes a new measurement every time it has traveled $0.2$ from its previous measurement, and an episode terminates when all agents exhaust their budget.
We use the Adam optimizer with a learning rate $5\times 10^{-5}$, which decays by a factor $0.96$ every $32$ steps, and train with batch size $512$.
PPO updates the neural network for $8$ iterations at each training step.
Our model is trained using one i9-10980XE CPU and one NVIDIA GeForce RTX 3090 GPU, and needs about 20 hours to fully converge.
Our code is available at \url{https://github.com/marmotlab/MAIPP}.

%%%%%%%%%%%%%%%%%%%%%%%%%%%%%%%%%%%%%%%%%%%%%%%%%%%%%%%%%%%%%%%%%%%%%%%%%%%%%%%%
%%%%%%%%%%%%%%%%%%%%%%%%%%%%%%%%%%%%%%%%%%%%%%%%%%%%%%%%%%%%%%%%%%%%%%%%%%%%%%%%

\section{EXPERIMENTS}
\label{Experiments}
\begin{figure}[t]
\centering
\subfigure[Destination Intent]{
\begin{minipage}[b]{0.45\linewidth}
\centering
\includegraphics[width=4.2cm, height=3.6cm]{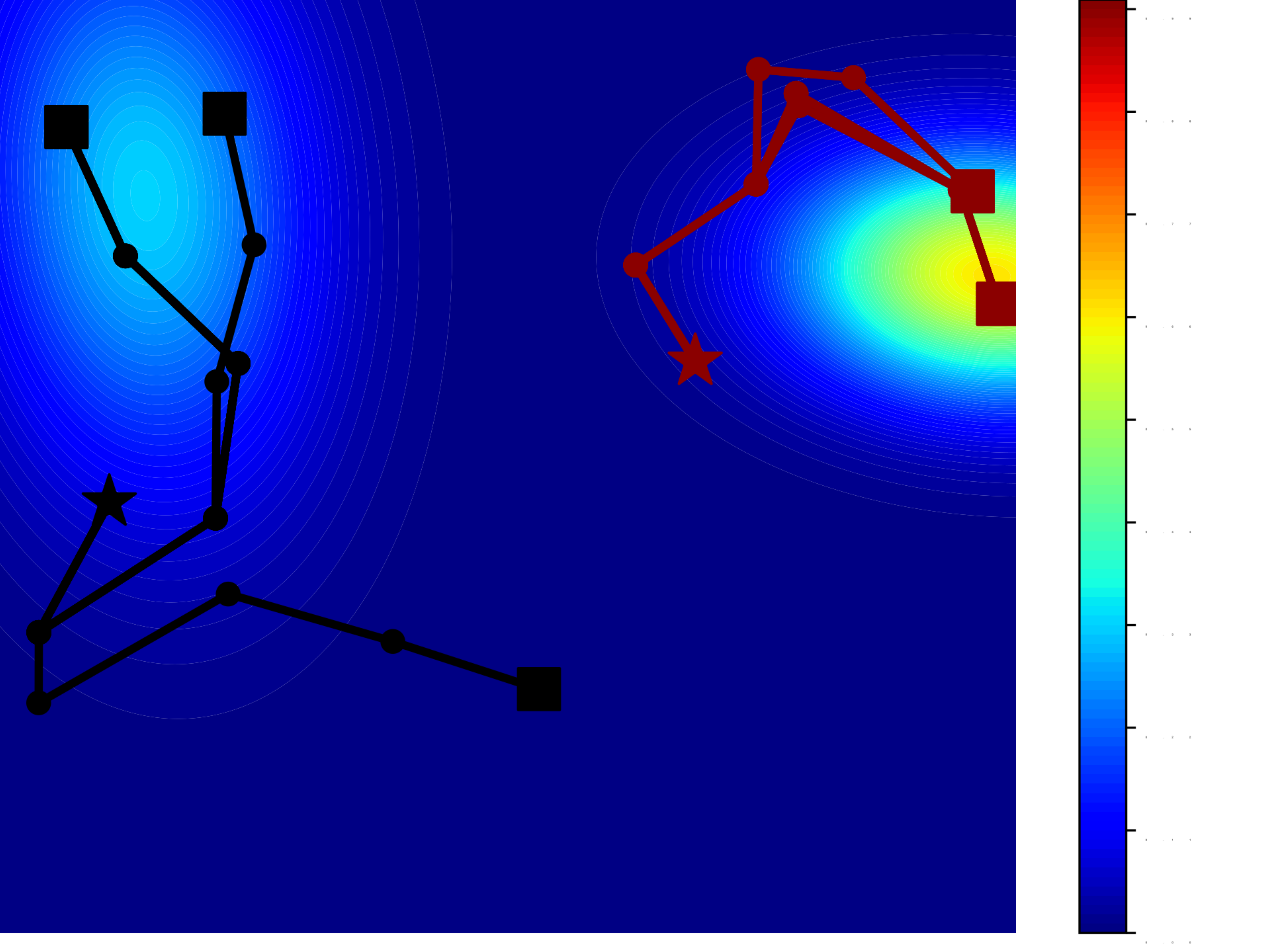}
\end{minipage}%
}%
\subfigure[Trajectory Intent]{
\begin{minipage}[b]{0.45\linewidth}
\centering
\includegraphics[width=4.2cm, height=3.6cm]{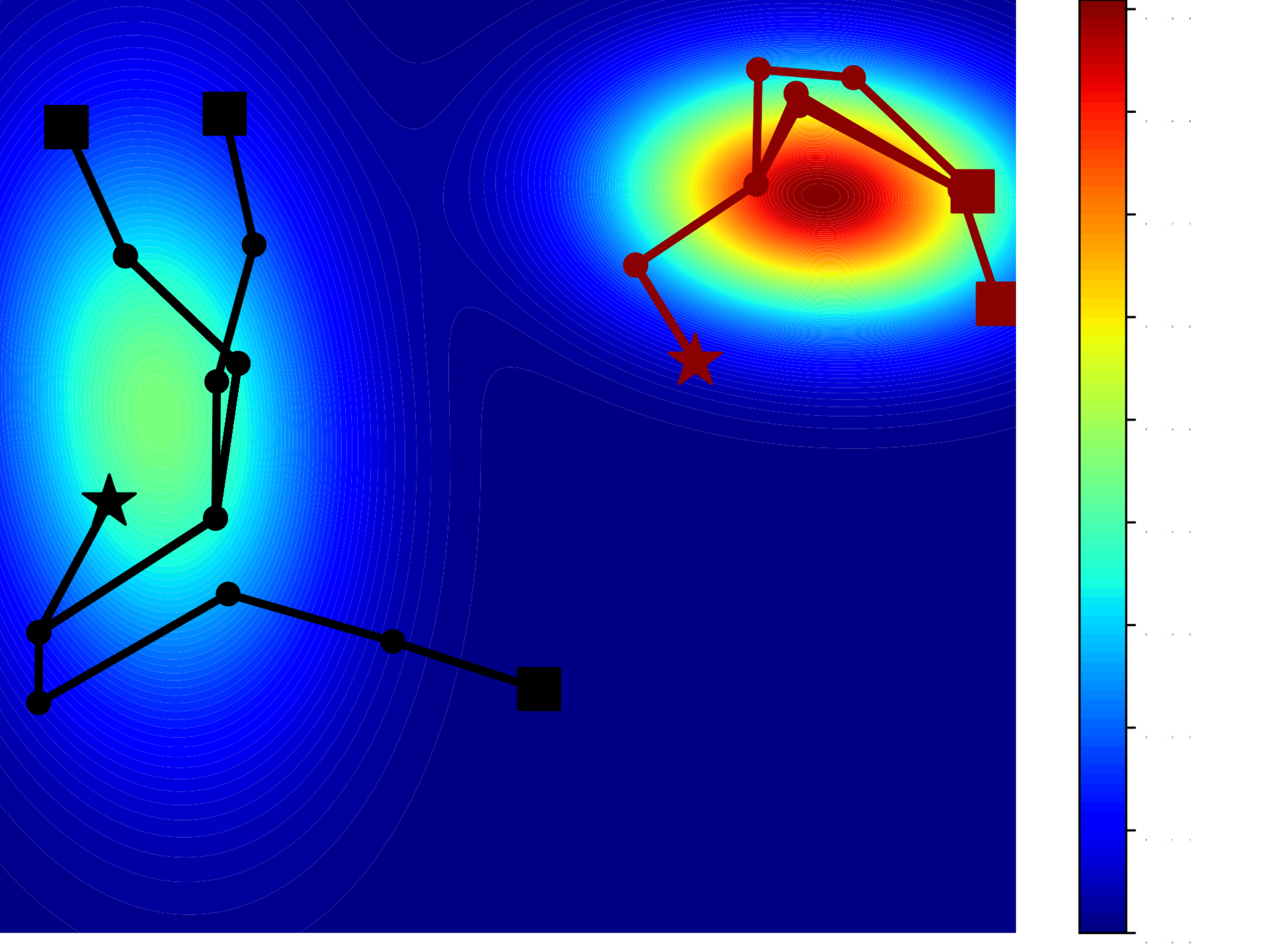}
\end{minipage}
}
\vspace{-0.2cm}
\caption{\textbf{Our two intent generation methods, for the same sampled trajectories (colored lines, here composed of $5$ waypoints each) from the agent's current position (star)}: Destination intent (DI) fits a Gaussian distribution over the final node of each trajectory only (colored squares), which can result in a wider Gaussian when these points are more spread out, whereas trajectory intent (TI) relies on all waypoints (colored circles and squares) along all trajectories, often resulting in a more peaked Gaussian since many of these nodes are more clumped together.}
\label{intent}
\vspace{-0.5cm}
\end{figure}

In our experiments, all agents also share the same start position $v_1$ and the same budget $B$. 
We further investigate two methods to let each agent $i$ aggregate its set of sampled predicted trajectories $\{g^i(t)\}$ into an intent distribution:

\begin{itemize}
    \item \textit{destination intent} (DI), where the intent distribution is built by fitting a Gaussian distribution (GD) over the set of all \textit{end nodes} of the sampled trajectories only, and

    \item \textit{trajectory intent} (TI), where the intent distribution is generated by fitting a GD over all nodes along all sampled trajectories. See Fig.~\ref{intent} for a visual comparison.
\end{itemize}

%%%%%%%%%%%%%%%%%%%%%%%%%%%%%%%%%%%%%%%%%%%%%%%%%%%%%%%%%%%%%%%%%%%%%%%%%%%%%%%%

\subsection{Results}

We compare our method with two sequential greedy assignment (SGA) based baselines: one using an underlying RRT~\cite{hollinger2014sampling} based single-agent IPP planner, while the other relies on the state-of-the-art single-agent DRL-based planner CAtNIPP~\cite{cao2022catnipp} (greedy variant, using the purely reactive model). 
In SGA+RRT, agents plan their individual paths sequentially, considering the current belief and paths selected by previous agents. Specifically, each agent samples numerous paths and finally selects the best one to be executed (i.e., the one that minimizes ${\rm Tr}(P_f)$). They execute a portion of the best path ($0.2$, in practice), update their belief, and then replan the next sub-path. SGA+CAtNIPP uses a reactive planner (CAtNIPP) where the agent only selects the next node to visit at each decision step.

Table~\ref{3 agents results} shows the comparison results between our methods and these SGA baselines. We vary the agents' budget $B$ (same for each agent) and present results for $m=3$ agents. RRT$(a,b)$ indicates that agents select the best path from all branches of the generated tree considering path lengths in the range $[a,b]$ (typically more than $8$ paths). For our DI/TI variants, $(a, j)$ means agents sample $a$ paths with $j$ nodes to generate their intent distribution. In TI$(8,5)^*$, agents select the first step of the best path sampled during intent generation as their next node to visit. We also extend the evaluation to $m=10$ agents in Table~\ref{10 agents results} by reducing the sensor range.

In many scenarios, such as disaster relief operations or ocean exploration, agents may need to accommodate limited communication ranges. We conducted additional experiments with fixed communication ranges of $0.3$ and $0.6$. Agents acquire beliefs and intent only from agents within the communication range during decision-making. The results are summarized in Table~\ref{partial communication}.

%%%%%%%%%%%%%%%%%%%%%%%%%%%%%%%%%%%%%%%%%%%%%%%%%%%%%%%%%%%%%%%%%%%%%%%%%%%%%%%%

\subsection{Discussion}

From our results in Table~\ref{3 agents results} and Table~\ref{10 agents results}, we first observe that our variants outperform both SGA+RRT and SGA+CAtNIPP across all experiments with varying budgets.
Second, we also observe that SGA+CAtNIPP consistently outperforms SGA+RRT.
We believe that the reason behind these performance differences is that the learning-based approaches (SGA+CAtNIPP and our variants) allow the agent to sequence reactive, non-myopic decisions based on its entire belief, whereas RRT may likely remain short-sighted since it can only rely on a finite sampling horizon.
Second, we note that SGA+CAtNIPP is equivalent to our method without the additional intent input, thus making this comparison an ablation study as well.
There, the fact that any of our variants outperforms SGA+CAtNIPP highlights the added performance obtained from allowing agents to learn to cooperate through these medium-/long-term predictions, even when those may be inaccurate (e.g., when longer paths are used to generate individual intent maps).

We then investigated how the performance of SGA+RRT is affected by the planning horizon, and results can be found in Table~\ref{3 agents results}.
There, we show that, unsurprisingly, SGA+RRT sees a sharp decline in performance as the planning horizon increases.
That is, as agents share longer paths, which they often do not execute as they replan along the way (likely to very different new paths), this type of path sharing is actually hurting the team performance by introducing noise/confusion in the sampling process. Therefore, the longer the sampling path, the less stable the predicted future intent.

\begin{table*}[t]
\vspace*{0.5cm}
\centering
\caption{\centering Comparison between our two sampling variants, in terms of ${\rm Tr}(P_f)$, for $3$ agents (10 trials on 30 instances).}
\vspace{-0.3cm}
\begin{tabular}{c|ccl|ccl|ccl|ccl|}
\hline
Method                  & \multicolumn{3}{c|}{\begin{tabular}[c]{@{}c@{}}Budget 2\\ for each agent\end{tabular}} & \multicolumn{3}{c|}{\begin{tabular}[c]{@{}c@{}}Budget 3\\ for each agent\end{tabular}} & \multicolumn{3}{c|}{\begin{tabular}[c]{@{}c@{}}Budget 4\\ for each agent\end{tabular}} & \multicolumn{3}{c}{\begin{tabular}[c]{@{}c@{}}Budget 5\\ for each agent\end{tabular}} \\ \hline
        & mean                                 & \multicolumn{2}{c|}{std}                        & mean                                 & \multicolumn{2}{c|}{std}                        & mean                                 & \multicolumn{2}{c|}{std}                        & mean                                & \multicolumn{2}{c}{std}                         \\ \hline
SGA+RRT(0.9, 1.0)           & 178.37                               & \multicolumn{2}{c|}{6.66}                       & 151.59                               & \multicolumn{2}{c|}{3.44}                       & 145.15                               & \multicolumn{2}{c|}{6.47}                       & 135.41                              & \multicolumn{2}{c}{6.74}                        \\
SGA+RRT(0.6, 0.7)           & 102.03                               & \multicolumn{2}{c|}{3.80}                       & 55.95                                & \multicolumn{2}{c|}{3.51}                       & 45.44                                & \multicolumn{2}{c|}{1.80}                       & 41.47                               & \multicolumn{2}{c}{3.21}                        \\
SGA+RRT(0.3, 0.4)           & 39.42                                & \multicolumn{2}{c|}{1.91}                       & 16.90                                & \multicolumn{2}{c|}{1.10}                       & 9.16                                 & \multicolumn{2}{c|}{0.65}                       & 5.46                                & \multicolumn{2}{c}{0.41}                        \\
SGA+CAtNIPP               & 38.75                                & \multicolumn{2}{c|}{3.05}                       & 13.39                                & \multicolumn{2}{c|}{0.91}                       & 5.61                                 & \multicolumn{2}{c|}{0.34}                       & 3.27                                & \multicolumn{2}{c}{0.29}                        \\
Destination Intent(8,3) & 37.43                                & \multicolumn{2}{c|}{1.27}                       & 11.55                                & \multicolumn{2}{c|}{0.89}                       & 4.56                                 & \multicolumn{2}{c|}{0.41}                       & 2.82                                & \multicolumn{2}{c}{0.11}                        \\
Trajectory Intent(8,5)  & 32.85                                & \multicolumn{2}{c|}{2.07}                       & 8.25                                 & \multicolumn{2}{c|}{0.61}                       & 3.66                                 & \multicolumn{2}{c|}{0.27}                       & 2.34                                & \multicolumn{2}{c}{0.10}                        \\
Trajectory Intent(8,5)* & \textbf{30.30}                       & \multicolumn{2}{c|}{1.26}                       & \textbf{7.76}                        & \multicolumn{2}{c|}{0.34}                       & \textbf{3.32}                        & \multicolumn{2}{c|}{0.24}                       & \textbf{1.99}                       & \multicolumn{2}{c}{0.16}                        \\ \hline

\end{tabular}
\label{3 agents results}
\vspace{-0.2cm}
\end{table*}

\begin{table}[t]
\vspace{-0.15cm}
\centering
\caption{\centering Performance comparison in terms of final average ${\rm Tr}(P_f)$, for $10$ agents (10 trials on 30 instances).}
% \resizebox{\linewidth}{!}{
\vspace{-0.3cm}
\begin{tabular}{c|ccl|ccl|ccl|}
\hline
Method          & \multicolumn{3}{c|}{\begin{tabular}[c]{@{}c@{}}Budget 2\\ for each agent\end{tabular}} & \multicolumn{3}{c|}{\begin{tabular}[c]{@{}c@{}}Budget 3\\ for each agent\end{tabular}} & \multicolumn{3}{c}{\begin{tabular}[c]{@{}c@{}}Budget 4\\ for each agent\end{tabular}} \\ \hline
different graph & mean                                 & \multicolumn{2}{c|}{std}                        & mean                                 & \multicolumn{2}{c|}{std}                        & mean                                & \multicolumn{2}{c}{std}                         \\ \hline
SGA+RRT             & 30.50                                & \multicolumn{2}{c|}{1.79}                       & 16.12                                & \multicolumn{2}{c|}{1.33}                       & 10.85                               & \multicolumn{2}{c}{0.88}                        \\ 
SGA+CAtNIPP              & 27.83                                & \multicolumn{2}{c|}{3.45}                       & 8.90                                 & \multicolumn{2}{c|}{1.17}                       & 4.51                                & \multicolumn{2}{c}{0.30}                        \\
DI (8,3)         & 23.86                                & \multicolumn{2}{c|}{3.09}                       & 7.14                                 & \multicolumn{2}{c|}{0.39}                       & 3.81                                & \multicolumn{2}{c}{0.44}                        \\
TI (8,5)*        & \textbf{19.39}                       & \multicolumn{2}{c|}{2.34}                       & \textbf{5.87}                        & \multicolumn{2}{c|}{0.46}                       & \textbf{3.07}                       & \multicolumn{2}{c}{0.34}                        \\ \hline

\end{tabular}
\label{10 agents results}
% \vspace{-0.5cm}
\end{table}

\begin{table}[t]
\centering
\caption{\centering Average final ${\rm Tr}(P_f)$ for different numbers of sampled waypoints during the individual intent generation in DI/TI.}
% \resizebox{\linewidth}{!}{
\vspace{-0.3cm}
\begin{tabular}{c|ccl|ccl}
\hline
\multirow{2}{*}{}       & \multicolumn{3}{c|}{\begin{tabular}[c]{@{}c@{}}Budget 3\\ for each agent\end{tabular}} & \multicolumn{3}{c}{\begin{tabular}[c]{@{}c@{}}Budget 4\\ for each agent\end{tabular}} \\ \cline{2-7} 
                        & mean                                 & \multicolumn{2}{c|}{std}                        & mean                                 & \multicolumn{2}{c}{std}                        \\ \hline
DI (8,3) & \textbf{11.55}                       & \multicolumn{2}{c|}{0.89}                       & \textbf{4.56}                        & \multicolumn{2}{c}{0.41}                       \\
DI (8,6) & 12.53                                & \multicolumn{2}{c|}{0.45}                       & 4.81                                 & \multicolumn{2}{c}{0.39}                       \\
DI (8,9) & 12.61                                & \multicolumn{2}{c|}{0.91}                       & 5.24                                 & \multicolumn{2}{c}{0.45}                       \\ \hline
TI (8,3)  & 9.65                                 & \multicolumn{2}{c|}{0.72}                       & 3.94                                 & \multicolumn{2}{c}{0.25}                       \\
TI (8,6)  & \textbf{8.43}                        & \multicolumn{2}{c|}{0.48}                       & \textbf{3.67}                        & \multicolumn{2}{c}{0.21}                       \\
TI (8,9)  & 8.92                                 & \multicolumn{2}{c|}{0.78}                       & 3.91                                 & \multicolumn{2}{c}{0.30}                       \\ \hline
\end{tabular}
\label{sampling steps with cov_trace}
\vspace{-0.5cm}
\end{table}

\begin{table}[t]
\vspace{-0.15cm}
\centering
\caption{\centering Performance under limited communication ranges, in terms of ${\rm Tr}(P_f)$, for $3$ agents (10 trials on 30 instances)}
\vspace{-0.4cm}
\begin{tabular}{c|ccl|ccl|ccl}
\hline
                                                                     & \multicolumn{3}{c|}{Budget 2}                 & \multicolumn{3}{c|}{Budget 3}                 & \multicolumn{3}{c}{Budget 4}                \\ \hline
\begin{tabular}[c]{@{}c@{}}Communication \\ Range = 0.3\end{tabular} & mean              & \multicolumn{2}{c|}{std}  & mean              & \multicolumn{2}{c|}{std}  & mean             & \multicolumn{2}{c}{std}  \\ \hline
SGA+RRT(0.3,0.4)                                                     & 99.53             & \multicolumn{2}{c|}{6.33} & 52.29             & \multicolumn{2}{c|}{3.95} & 26.82            & \multicolumn{2}{c}{3.61} \\
SGA+CAtNIPP                                                          & 65.62             & \multicolumn{2}{c|}{4.59} & 25.35             & \multicolumn{2}{c|}{2.24} & 11.11            & \multicolumn{2}{c}{0.80} \\
DI (8,3)                                                             & 66.59             & \multicolumn{2}{c|}{3.57} & 24.34             & \multicolumn{2}{c|}{1.99} & 9.74             & \multicolumn{2}{c}{0.59} \\
TI (8,5)*                                                            & \textbf{50.95}    & \multicolumn{2}{c|}{2.34} & \textbf{16.04}    & \multicolumn{2}{c|}{1.82} & \textbf{6.21}    & \multicolumn{2}{c}{0.27} \\ \hline
\begin{tabular}[c]{@{}c@{}}Communication \\ Range = 0.6\end{tabular} & mean              & \multicolumn{2}{c|}{std}  & mean              & \multicolumn{2}{c|}{std}  & mean             & \multicolumn{2}{c}{std}  \\ \hline
SGA+RRT(0.3,0.4)                                                     & 54.96             & \multicolumn{2}{c|}{6.31} & 26.75             & \multicolumn{2}{c|}{1.62} & 17.86            & \multicolumn{2}{c}{2.17} \\
SGA+CAtNIPP                                                          & 54.02             & \multicolumn{2}{c|}{2.91} & 20.16             & \multicolumn{2}{c|}{2.54} & 8.87             & \multicolumn{2}{c}{0.75} \\
DI(8,3)                                                              & 51.70             & \multicolumn{2}{c|}{3.72} & 18.00             & \multicolumn{2}{c|}{1.73} & 7.61             & \multicolumn{2}{c}{1.07} \\
TI(8,5)*                                                             & \textbf{40.99}    & \multicolumn{2}{c|}{3.07} & \textbf{11.39}    & \multicolumn{2}{c|}{0.65} & \textbf{5.12}    & \multicolumn{2}{c}{0.65} \\ \hline
\begin{tabular}[c]{@{}c@{}}Global \\ Communication\end{tabular}      & mean              & \multicolumn{2}{c|}{std}  & mean              & \multicolumn{2}{c|}{std}  & mean             & \multicolumn{2}{c}{std}  \\ \hline
TI(8,5)*                                                             & 30.30             & \multicolumn{2}{c|}{1.26} & 7.76              & \multicolumn{2}{c|}{0.34} & 3.32             & \multicolumn{2}{c}{0.24} \\ \hline
\end{tabular}
\label{partial communication}
\vspace{-0.6cm}
\end{table}

However, when we look at the average performance (final ${\rm Tr}(P_f)$, reported in Table~\ref{sampling steps with cov_trace}) of these variants, the story is very different.
There, for our DI variant, we observe the same correlation between prediction stability 
and final performance as for the SGA+RRT baseline.
That is, DI seems to perform best when agents share shorter-term, more stable predicted intents.
However, and most interestingly, our TI variant seems to break free from this trend: its best performance is found when agents share intent maps created using medium-length trajectories (composed of $6$ nodes in our case), despite these maps exhibiting lower stability in time.
We believe that this confirms our hypothesis, about the attention-based structure of our neural network allowing agents to dynamically learn which portion(s) of the shared intent maps to trust/rely on to maximize their performance, even/especially in cases where these intent may not be fully accurate.
That is, agents in our TI variant are able to extract more information from intent maps constructed from medium-length sampled trajectories, which are inherently less stable/accurate, leading to improved overall performance.
That being said, we still note that this performance comes back down if the length of the sampled trajectories is further increased to $9$ nodes, where the accuracy/stability of the intent map decreases below some critical level, introducing too much noise/confusion in the team and leading to worse final performances.

Table~\ref{partial communication} shows that traditional methods like RRT-based approaches have high communication requirements. When the communication range is small (e.g., $0.3$), their performance significantly drops. However, increasing the range to 0.6 improves their performance, bringing it close to the global communication scenario. This is because, without acquiring the beliefs of other agents, the SGA+RRT method allows agents to select paths solely based on their own belief, resulting in wasted budget from redundant work. Even with a large budget, SGA+RRT performs poorly as agents lack accurate enough information to distribute well and cooperatively cover the domain.

In contrast, learning-based methods such as SGA+CatNIPP and our intent-based method excel at making non-myopic decisions even under limited communication.
Although we notice a slight decrease in performance ($\sim 20 \%$) between performances under global and 0.6 communication ranges, the learning-based methods do not experience a significant decline like SGA+RRT.
This gap between traditional and learning-based approaches is further magnified when the communication range is decreased to 0.3.
We believe that these results highlight how learning-based approaches enable agents to identify relevant/accurate portions of the incomplete/inaccurate information available to them, particularly through the use of attention mechanisms, leading to improved decision-making.
More importantly, we note that our model trained with global communications keeps promising performance even under limited communication ranges, showcasing its generalizability. We believe that training models specifically under known communication constraints in real-world applications could yield even better performance.

%%%%%%%%%%%%%%%%%%%%%%%%%%%%%%%%%%%%%%%%%%%%%%%%%%%%%%%%%%%%%%%%%%%%%%%%%%%%%%%%
%%%%%%%%%%%%%%%%%%%%%%%%%%%%%%%%%%%%%%%%%%%%%%%%%%%%%%%%%%%%%%%%%%%%%%%%%%%%%%%%

\vfill
\section{CONCLUSION}
\label{conclusion}
This paper introduces a decentralized, policy-based deep reinforcement learning model for adaptive MAIPP, where each agent utilizes its own policy network to predict its medium-/long-term behavior, and shares this information with others to enhance cooperative decision-making. We propose two methods to aggregate shared information into individual decision-making and compare their performance. Our approach demonstrates advantages across various budgets and team sizes without any retraining, where it outperforms two SGA-based MAIPP baselines: a conventional single-agent IPP solver, and a state-of-the-art learning-based approach. Our experiments indicate that the improved performance of our approach is likely due to learned attention mechanisms, enabling agents to determine which portions of shared intent they can trust and rely on, even in cases of inaccuracy or rapidly changing information. Additionally, we present results under limited communication ranges, demonstrating that our trained model maintains significant performance improvements in realistic constraint scenarios not encountered during training.

We emphasize that our approach to generating, sharing, and utilizing intent for cooperative decision-making extends beyond MAIPP and can be applied or extended to other multi-agent tasks, such as multi-agent pathfinding, task allocation, or collective robotic construction. Future work will explore these extensions and applications, as well as implement our models on hardware, leveraging the reactive nature of our approach to meet real-time deployment requirements.

%%%%%%%%%%%%%%%%%%%%%%%%%%%%%%%%%%%%%%%%%%%%%%%%%%%%%%%%%%%%%%%%%%%%%%%%%%%%%%%%
%%%%%%%%%%%%%%%%%%%%%%%%%%%%%%%%%%%%%%%%%%%%%%%%%%%%%%%%%%%%%%%%%%%%%%%%%%%

\newpage
\section*{ACKNOWLEDGMENT}

This work was supported by Temasek Laboratories (TL@NUS) under grants TL/SRP/21/19 and TL/FS/2022/01.

%%%%%%%%%%%%%%%%%%%%%%%%%%%%%%%%%%%%%%%%%%%%%%%%%%%%%%%%%%%%%%%%%%%%%%%%%%%%%%%%
%%%%%%%%%%%%%%%%%%%%%%%%%%%%%%%%%%%%%%%%%%%%%%%%%%%%%%%%%%%%%%%%%%%%%%%%%%%%%%%%

\bibliographystyle{IEEEtran} 
\bibliography{ref.bib}

\end{document}